\documentclass[]{article}
\usepackage[english]{babel}

\usepackage[utf8]{inputenc}
\usepackage{times}
\usepackage{graphicx} 

\usepackage{float}
\usepackage{natbib}
\usepackage{booktabs}

\usepackage{amsmath,amsthm, amssymb}
\usepackage{xcolor}
\usepackage{subcaption}
\usepackage[acronym]{glossaries}
\usepackage{microtype}

\usepackage[switch]{lineno}

\usepackage[accepted]{icml2017}

\newacronym{lstm}{LSTM}{Long Short Term Memory}
\newacronym{rnn}{RNN}{Recurrent Neural Network}
\newacronym{nnlm}{NNLM}{Neural Network Language Models}
\newacronym{lm}{LM}{Language Modelling}
\newacronym{nlp}{NLP}{Natural Language Processing}
\newacronym{lsa}{LSA}{Latent Semantic Analysis}
\newacronym{vsm}{VSM}{Vector Space Models}
\newacronym{ir}{IR}{Information Retrieval}
\newacronym{hal}{HAL}{Hyperspace Analogue to Language}
\newacronym{svd}{SVD}{Singular Value Decomposition}
\newacronym{bnc}{BNC}{British National Corpus}
\newacronym{bow}{BoW}{Bag-of-Words}
\newacronym{jl}{JL}{Johnson-Lindenstrauss}
\newacronym{sdm}{SDM}{Sparse Distributed Memory}
\newacronym{nrp}{NRP}{Neural Random Projection}
\newacronym{nplm}{NPLMs}{Neural Probabilistic Language Models}
\newacronym{som}{SOM}{Self-Organising Map}
\newacronym{nce}{NCE}{Noise Constrastive Estimation}
\newacronym{sgd}{SGD}{Stochastic Gradient Descent}

\icmltitlerunning{Neural Random Projections}

\begin{document}
	
	\twocolumn[
	\icmltitle{Neural Random Projections for Language Modelling}
	
	%

\begin{icmlauthorlist}
	\icmlauthor{Davide Nunes}{ul}
	\icmlauthor{Luis Antunes}{ul}
	
\end{icmlauthorlist}

\icmlaffiliation{ul}{BioISI---Instituto de Biosistemas e Ciências Integrativas,\\ Faculdade de 
Ciências, Universidade de Lisboa, Portugal.}

\icmlcorrespondingauthor{Davide Nunes}{davex@ciencias.ulisboa.pt}
\icmlcorrespondingauthor{Luis Antunes}{xarax@ciencias.ulisboa.pt}

\paperkeywords{neural networks, random projections, online learning, distributional semantics, 
	natural language processing, semantics, computational semantics, language modelling}
\vskip 0.3in
]
\printAffiliations{}

\begin{abstract}
\noindent Neural network-based language models deal with data sparsity problems by mapping the 
large discrete space of words into a smaller continuous space of real-valued vectors. By learning 
distributed vector representations for words, each training sample informs the neural network model 
about a combinatorial number of other patterns. In this paper, we exploit the sparsity in natural 
language even further by encoding each unique input word using a fixed sparse random representation. 
These sparse codes are then projected onto a smaller embedding space which allows for the encoding 
of word occurrences from a possibly unknown vocabulary, along with the creation of more compact 
language models using a reduced number of parameters. We investigate the properties of our encoding 
mechanism empirically, by evaluating its performance on the widely used Penn Treebank corpus. 
We show that guaranteeing approximately equidistant (nearly orthogonal) vector representations for unique 
discrete inputs is enough to provide the neural network model with enough information to learn --and make 
use-- of distributed representations for these inputs.

\end{abstract}

\section{Introduction}
\label{sec:intro}
The goal of computational semantics is to automate the learning of meaningful representations for 
natural language expressions. In particular, representation learning is integral to 
state-of-the-art \acrfull{lm}, which is a keystone in computational semantics 
used for a wide range of \acrfull{nlp} tasks, from information retrieval \cite{Ponte1998} 
to speech recognition \cite{Arisoy2012}, or machine translation~\cite{Koehn2010}. Better language 
models frequently lead to improved performance on underlying downstream tasks, which makes 
\acrshort{lm} valuable in itself. Moreover, certain domains allow language models to extract 
knowledge implicitly encoded in the training data. For example, in \cite{Serban2015}, learning from 
film subtitles allows models to answer basic questions about colours and people. Learning 
representations for discrete symbols is particularly useful in the context of language models 
because capturing regularities in relationships between words (which are not only complex but 
recursive in nature) allows us to build better models. The \acrshort{lm}, as a task, can 
thus be seen as a framework for the study of more general learning approaches operating in fairly 
complex and sometimes dynamic scenarios.

The goal of \acrshort{lm} is to model the joint probability distribution of words in a text corpus. 
Models are trained by maximising the probability of a word given all the previous words in 
the training dataset, formally:
\begin{equation*}
P(w_1,\ldots,w_{N}) = \prod_{i=1}^{N} P(w_i|h_i),
\end{equation*}
\noindent where $w_i$ is the current word and $c_i=(w_1,\cdots,w_{i-1})$ is the current word 
\textit{history} or \textit{context}. A Markov assumption is usually used as approximation: instead 
of using the full word history as context, the probability of observing the $i^{th}$ 
word $w_i$ is approximated by the probability of observing it in a shortened context of $n - 1$ 
preceding words. To train a language model under this assumption, a \textit{sliding window} 
$(w_i,c_i)$ of size $n$ is applied sequentially across the textual data. Each window is also 
commonly referred to as \textit{n-gram}. 

Traditional \textit{n-gram} models \cite{Brown1992,Kneser1995} are learned by building conditional 
probability tables for each $P(w_i|c_i)$. This approach is popular due to its simplicity and 
overall good performance, but this formulation poses some obstacles. 
More specifically, we can see that even with large training corpora, extremely small or zero 
probabilities can be given to valid sequences, mostly due to rare word occurrences. Dealing with 
this sparseness and unseen data requires smoothing techniques that reallocate probability mass 
from observed to unobserved \textit{n-grams}, producing better estimates for unseen data 
\cite{Kneser1995}. The fundamental issue is that classic \textit{n-gram} models lack a notion of 
word similarity. Words are treated as equality likely to occur discrete entities with no relation 
to each other. This makes density estimation inherently difficult, as there is no straightforward 
way to perform smoothing. As an example, there is no way to estimate the probability for two sequences differing only in a synonymic expression. 

Neural networks have been used as a way to deal with both the sparseness and smoothing problems. 
The work in \cite{Bengio2003} represents a paradigm shift for language modelling and an example of 
what we call \acrfull{nnlm}. In a \acrshort{nnlm}, the probability distribution for a word given 
its context is modelled as a smooth function of learned real-valued vector representations for each 
word in that context. The neural network models learn a conditional 
probability function $P(w_i|c_i;\theta_F,\theta_L)$, where $\theta_F$ is a set of parameters 
representing the vector representations for each word and $\theta_L$ parameterises the 
log-probabilities of each word based on those learned features. A softmax function is then applied 
to these log-probabilities to produce a categorical probability distribution over the next word 
given its context. (In section \ref{sec:model}, we will present two instances of neural network models following this formulation.)

The resulting probability estimates are smooth functions of the continuous word vector 
representations, and so, a small change in such vector representations results in a small change in 
the probability estimation. Smoothing is learned implicitly, leading \acrshort{nnlm} to 
achieve better generalisation for unseen contexts. The resulting word vector representations are 
also commonly referred to as \textit{embeddings}, since they represent a mapping from the large 
discrete space (a vocabulary) to a smaller continuous space of real-valued vectors where they are 
``embedded". This constitutes a geometric analogy for meaning with the  that, with 
proper training, words that are semantically or grammatically related will be mapped to similar 
locations in the continuous space. 

Energy-based models provide a different perspective on statistical language modelling with 
parametric neural networks as a way to estimate joint discrete probability densities. Energy-based 
models such as \cite{Mnih2007,Mnih2009} capture dependencies between variables by 
associating a scalar energy score to each variable configuration. In this case, making predictions 
consists on setting the values for observed or visible variables and finding the value for the 
remaining (hidden) variables that minimise this energy score.

The key difference between the energy-based models in \cite{Mnih2007,Mnih2009} and the 
\acrshort{nnlm} in \cite{Bengio2003} is that instead of estimating a categorical distribution based 
on higher-level features produced by the \acrshort{nnlm} hidden layer based on the distributed 
representations for the context words, an energy-based model tries to predict the distributed 
representation of a target word and attributes an energy score to the output configuration of the 
model based on how close the prediction is from the actual representation for the target word. The 
correct representation for the target word is dependent on the current state of the model embedding 
parameters, but two separate parameter matrices can also be used for context and target words 
\cite{Mnih2012}. Most probabilistic models can in fact be viewed as a special type of energy-based model
in which the energy function has to satisfy certain normalisation conditions, along with a loss function 
with a particular form. Moreover, energy-based learning provides a unified framework for \textit{learning}, 
and can be seen as an alternative to probabilistic estimation for prediction or classification tasks 
\cite{Lecun2006}. 

In this paper, we describe an encoding mechanism that can be used with neural network 
probabilistic language models to represent unique discrete inputs while using a reduced parameter 
space. Our models use random projections as a way to encode input patterns, using a fixed size 
sparse random vector. The idea is to maintain the expressive power of the neural networks while 
reducing the parameter space and expanding the types of discrete patterns the network can use to 
make predictions. Following the principles from \cite{Achlioptas2003} and \cite{Kanerva2009}, our 
random projection encoding reduces the input dimension $|V|$ to a lower $k\text{-dimensional}$ 
dimensional space while approximately preserving the distances between expected input patterns. 
This means that instead of using the $1\text{-of-}V$ orthogonal encoding, we use a fixed 
$k\text{-dimensional}$ sparse vector with each input word being nearly-orthogonal to any other 
words. Our main hypothesis is that in order for a neural network to learn distributed 
representations for words, we do not require one unique vector representation per word; 
instead, we just need the representations of each pair of unique patterns to be approximately orthogonal. 
Each unique pattern should (by itself) be indistinguishable from any other unique 
pattern. By using random projections, we guarantee that each new representation is equally probable 
and nearly-orthogonal to any other representation. This encapsulates the idea that what makes a word 
unique in language is not its representation, but the way in which we use the word, since language 
is essentially a system of differences, where meaning in words arises through social social 
interaction \cite{Feyerabend1955}, and biological biases \cite{Pinker2010}.

\section{Background and Related Work}
\label{sec:background}
In this paper, the goal is two fold: to reduce the dimensionality of the input space, and to 
allow neural-network models to capture patterns that are usually intractable to represent 
using the typical count-based n-gram models. In the case of neural language modelling, each 
word is usually represented by a $1\text{-of-}V$ vector. The input space is an orthogonal 
unit matrix where each word is represented by a single value $1$ on the column corresponding 
to its index. This means that if we want to represent a more complex structured space such as
syntactic dependency trees, this would require either to use structured neural network 
\cite{Socher2011}, or to index all possible unique combinations of syntactic structures which
is intractable for large datasets. This article is a first step in the exploration the
possibility of encoding complex structured discrete spaces. We take an incremental and
bottom-up approach to the problem by exploring how to build an encoder that can be used in
word-level language models.

One cornerstone of neural-network-based models is the notion of distributed representation. This is, the fact that neural network can learn vector representations for discrete inputs as weights that are learned during the training process. In a distributed representation, each "thing" is represented as a combination of multiple factors. Learning distributed representations of concepts as patterns of activity in a neural network has been object of study in \cite{Hinton1986}. The mapping between unique concept ids and respective vector representations has been later referred to as \textit{embeddings}. The idea of using embeddings in language modelling is explored in the early work or Bengio et al. \cite{Bengio2003} and later popularised by the \textit{word2vec} method for word representation learning \cite{Mikolov2013a}. Both these methods encode input words using a $1\text{-of-}V$ encoding scheme, where each word is represented by its unique index. The size of the embedding space is thus proportional to the size of the vocabulary.

In the case of language modelling, there have been multiple developments to deal with large or unknown vocabularies. One straightforward method is to consider character or subword-level (e.g. morphemes) modelling instead of considering words the discrete input units \cite{Ling2015,Kim2016,Zhang2015}. The main motivation behind using subword-based modelling is that training accurate character-level models is extremely difficult and often more computationally expensive than working with models that use word-level inputs \cite{Mikolov2012}, but Elman shows that most of the entropy if concentrated at the first few characters of each word \cite{Elman1990}. Moreover, word-based models often fail to capture regularities in many inflectional and agglutinative languages. In this paper we focus on word-based modelling but the proposed encoding mechanism can be incorporated in models that use morphemes instead of words as inputs.

Despite the good results obtained with word-level neural probabilistic language models
\cite{Bengio2003,Mnih2007}, these have notoriously long training times, even for 
moderately-sized datasets. This is due to the fact that in order for these models to output
probability distributions, they require explicit normalisation, which requires us to consider
all words in the vocabulary to compute log-likelihood gradients. Solutions to this problem 
include structuring the vocabulary into a tree, which speeds up word probability computations
--but ties model predictive performance to the tree used \cite{Morin2005}. Another approach 
is to use a sampling approach to approximate gradient computations. In particular, 
\acrfull{nce} \cite{Gutmann2012} has been shown to be a stable approach to speed up 
unnormalised language models \cite{Mnih2012}. While we use full normalisation to analyse the 
predictive capability of our proposed models, much like in the previous case, the 
computational complexity introduced by the normalisation requirements can be alleviated using
techniques like \acrshort{nce}. Nevertheless, using random projections to model unique 
discrete inputs with neural networks, changes the dynamics of the language modelling problem.
This warrants an assessment of how effective approximation methods are --which we will do in 
future work.

Exploiting low-dimensional structure in high-dimensional problems has become a highly active 
area of research in machine learning, signal processing, and statistics. In summary, the goal
is to use a low-dimensional model of relevant data in order to achieve, better prediction, 
compression, or estimation compared to more complex "black-box" approaches to deal with 
high-dimensional spaces \cite{Hegde2016}. In particular, we are interested in the compression
aspect of these approaches. Usually, exploiting low-dimensional structure comes at a cost, 
which is the fact that incorporating structural constraints into a statistical estimation 
procedure often results in a challenging algorithmic problem. Neural network models have been
notoriously successful at dealing with these kind of constraints, and even at finding 
geometric analogies for complex patterns such as syntactic relationships \cite{Mikolov2013b}.
But the use of neural networks has been limited in the types of patterns that it can process 
and encode. Particularly, there is no straightforward way to encode priors about discrete 
input qualifiers or to encode structure within the input space --although some work has been 
done in that direction \cite{Vaswani2017}.

In this article we show that neural network models can be used to tackle a challenging 
statistical estimation problem such as language modelling, while using an approximation 
method to encode a large number of discrete inputs --and possibly allowing for the additional
encoding of linguistic-based structural constraints. To achieve this, we turn to a 
simple yet efficient class of dimensionality reduction methods, random projections.

The use of random projections as a dimensionality reduction tool has been extensively studied
before, especially in the context of approximate matrix factorisation --the reader can refer 
to \cite{Halko2011} for a more in depth review of this technique. The basic idea of random 
projections as dimensionality reduction technique comes from the work of Johnson
and Lindenstrauss, in which it was shown that the pairwise distances among a collection of 
$N$ points in an Euclidean space are approximately maintained when the points are mapped 
randomly to an Euclidean space of dimension $O(\log N)$. In other words, random embeddings 
preserve Euclidean geometry \cite{Johnson1984}. In the case of language modelling, we assume 
that finding representations for words is a problem of geometric nature (see \cite{Levy2014})
that can be translated to a lower-dimensional space and solved there. 

Random projections have been used in multiple machine learning problems as a way to speed-up 
computationally expensive methods. One example is the use of random mappings for fast 
approximate nearest-neighbour search \cite{Indyk2001}. Random projections have also been 
applied to document retrieval or representation learning based on matrix factorisation. In 
\cite{Papadimitriou1998}, random projections are used as a first step for Latent Semantic 
Analysis (\acrshort{lsa}), which is essentially a matrix factorisation of word co-occurrence 
counts~\cite{Landauer1997}. This makes the matrix factorisation algorithm much more 
tractable, because the projections preserve pair-wise distances used by the algorithm to find
latent vector representations for documents -- representations that explain word frequency  
variance well. A similar approach is followed in~\cite{Kaski1998}: a random mapping 
projection is applied to document vectors to reduce their dimensionality; these vectors then 
serve as input to a \acrfull{som} in order to cluster the documents according to their 
similarity. This approach was motivated by experiments from \cite{Ritter1989}, where instead 
of performing document clustering based on word co-occurrence frequencies, the authors 
performed word clustering \textit{after} reducing the context vector dimensionality using 
random projections. In the later work of Kanerva, an incremental random projection method 
called random indexing, is used to gather co-occurrence patters in an incremental fashion 
\cite{Kanerva2000}. This was motivated by previous work on associative memory models 
\cite{Kanerva1988}, and the work in \cite{Achlioptas2003}, which showed that a random 
projection matrix can be constructed incrementally using a simple sparse ternary distribution
(while still respecting the bound-preserving requirements from the Johnson and Lindenstrauss 
lemma \cite{Johnson1984}).

Our proposal of using random projections as the encoding mechanism for neural network 
language models is also related to \textit{dictionary learning}. Dictionary learning consists
of modelling signals as sparse linear combinations of atoms selected from a learned 
dictionary \cite{Gribonval2015}.

In this paper, we focus on exploring simple feedforward neural network architectures for
multiple reasons. First, using incrementally complex models allows us to determine which
architectural components allow neural network models to deal with random-projection encoding.
The work from Mnih et al. in 
energy-based neural language models \cite{Mnih2007,Mnih2009} is a good starting point for our
baseline because the energy-based model gives us a way to translate random vector predictions
into word probability distributions. 

Recurrent architectures have been successfully used in language modelling and related tasks 
\cite{Jozefowicz2016} --with these successes often being attributed to their capacity (in 
principle) to capture unbound contexts (retaining long term dependencies and ensuring that 
information can be  propagated through many time steps). We remain cautious about their use 
--as a baseline-- for our proposed approach. In practice, it is not clear whether or not the 
capacity to capture long-term dependencies (in principle) is key to achieve good results in 
all tasks related to \acrshort{lm}. For example, neural networks with a simple architecture 
and convolution operations, capable of dealing with \textit{fixed} contexts, have been 
proposed as a competitive alternative to \acrfull{lstm} networks \cite{Dauphin2016}. As 
another example, good performance can be achieve in a translation task 
by using attention mechanism with a feedforward architecture \cite{Vaswani2017}. 

Most importantly, using over-parametrised/complex architectures undermines our ability to 
distinguish between a model that performs well with a new component, and a model that performs 
well despite the new component. We do intend to explore the proposed encoding with recurrent 
architectures in the future, but for now, this is beyond the scope of this paper.

\section{NRP Model}
\label{sec:model}
In this section, we describe our neural random projection approach to language modelling. The 
main idea is to use a neural network model to learn lower-dimensional representations for discrete 
inputs while learning a discrete probability distribution of words similarly to \cite{Bengio2003, 
Mnih2007}. By contrast with the existing work, instead of encoding a word using a 
$1\text{-of-}V$ vector, each word is encoded using a random sparse high-dimensional vector
\cite{Kanerva2000,Kanerva2009}. This allows us to create neural-network models with a smaller number of parameters, 
but crucially, it also allows for more flexible patterns to be represented and for a model to be trained with a 
vocabulary of unknown size. 

\subsection{Random Input Encoding}
The distributed representations or \textit{embeddings} in our model are $k \times m$ subspace, with $k<<|V|$ 
(we have less base vectors than words in the known dictionary). A representation for each word as learned by 
the neural network is the linear combination of $(s \times m)\text{-dimensional}$ basis where $s$ is the number 
of non-zero entries for each vector. Each unique word is assigned a randomly generated sparse ternary vector 
we refer to as \emph{random index vector}. Random indices are sampled from the following distribution with 
mean $0$ and variance $1$:

\begin{equation}
    r_{i} = \left\{
    \begin{array}{rl}
    +1 & \mbox{with probability } \alpha/2 \\
    0 & \mbox{with probability } 1 - \alpha \hspace{1em}\\
    -1 & \mbox{with probability } \alpha/2
\end{array}
\right.
\label{eq:ri}
\end{equation}

\noindent where $\alpha$ is the proportion of non-zero entries in the random index vector. While similar to the
distribution proposed in \cite{Achlioptas2003}, we use very sparse random vectors and no scaling factor. Since the
input space is an orthogonal unit matrix, the inner product between any two random vectors $(r_i, r_j)$ is expected
to be concentrated at $0$. The motivation for this is to make any discrete input indistinguishable from any other
input in terms of distance between them. Instead of referring the sparseness parameter $a$, we will use $s$ as the
number of non-zero entries for randomly generated index vectors. This means that for $s=2$ we have exactly one entry
with value $1$, one with value $-1$ and all the rest have value $0$.

\begin{figure}[h]
\vskip 0.2in
	\begin{center}
		\centerline{\includegraphics[width=\columnwidth]{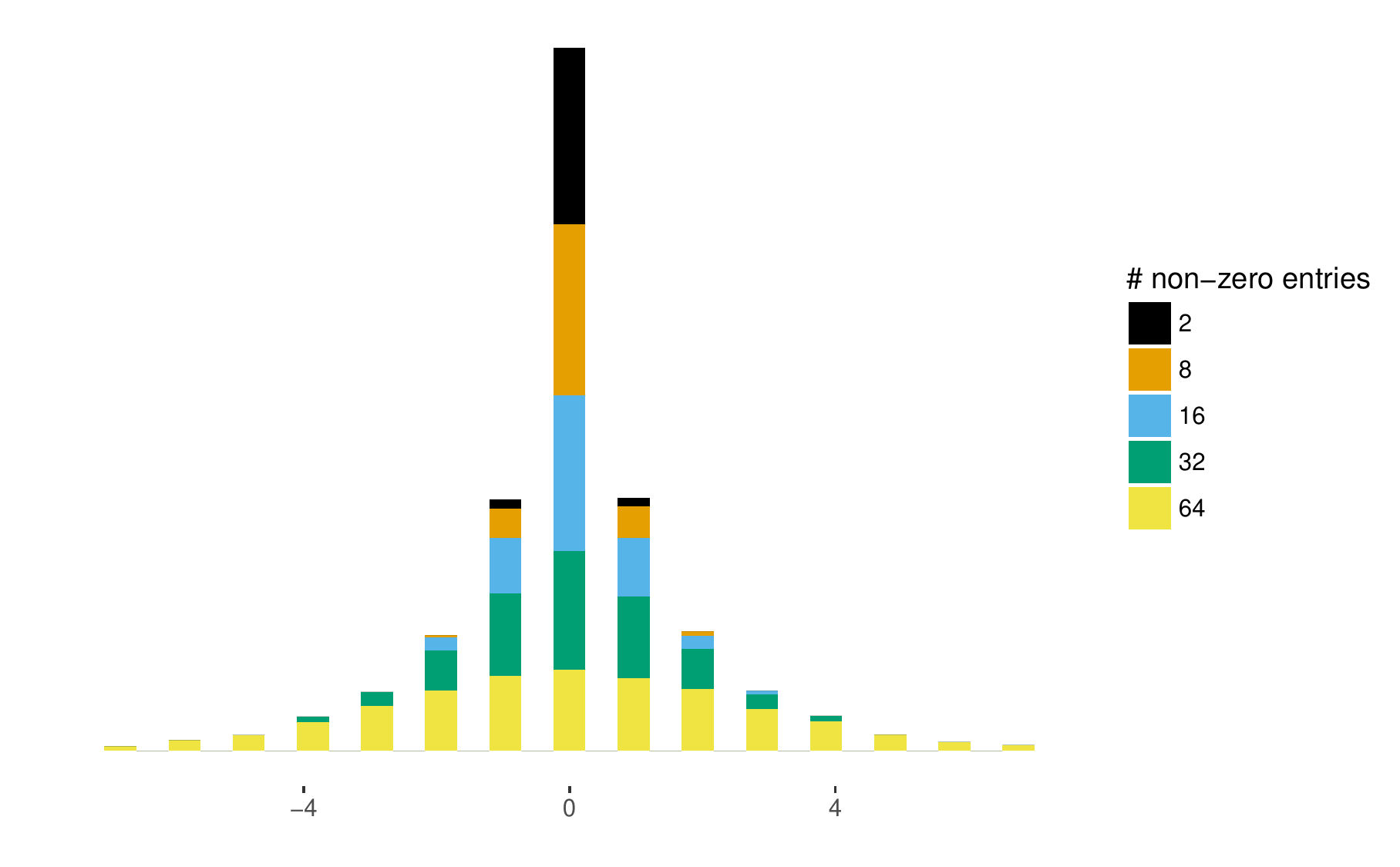}}
		\caption{Distribution of inner products between pairs of randomly generated 
		1000-dimensional random 
		indices.}
		\label{fig:ri_1000}
	\end{center}
	\vskip -0.2in	
\end{figure} 

Figure {\ref{fig:ri_1000}} shows the distribution of inner products for a $1000-k$ dimensional random 
indices. A larger number of non-zero entries leads to a greater probability of a collision to occur, 
but the inner products remain well concentrated at $0$ -- any two random sparse vectors are expected 
to be either orthogonal or near orthogonal.

\begin{figure*}
\begin{subfigure}[b]{1\columnwidth}
\vskip 0.1in
\centering
	\begin{center}
		\centerline{\includegraphics[width=\linewidth]{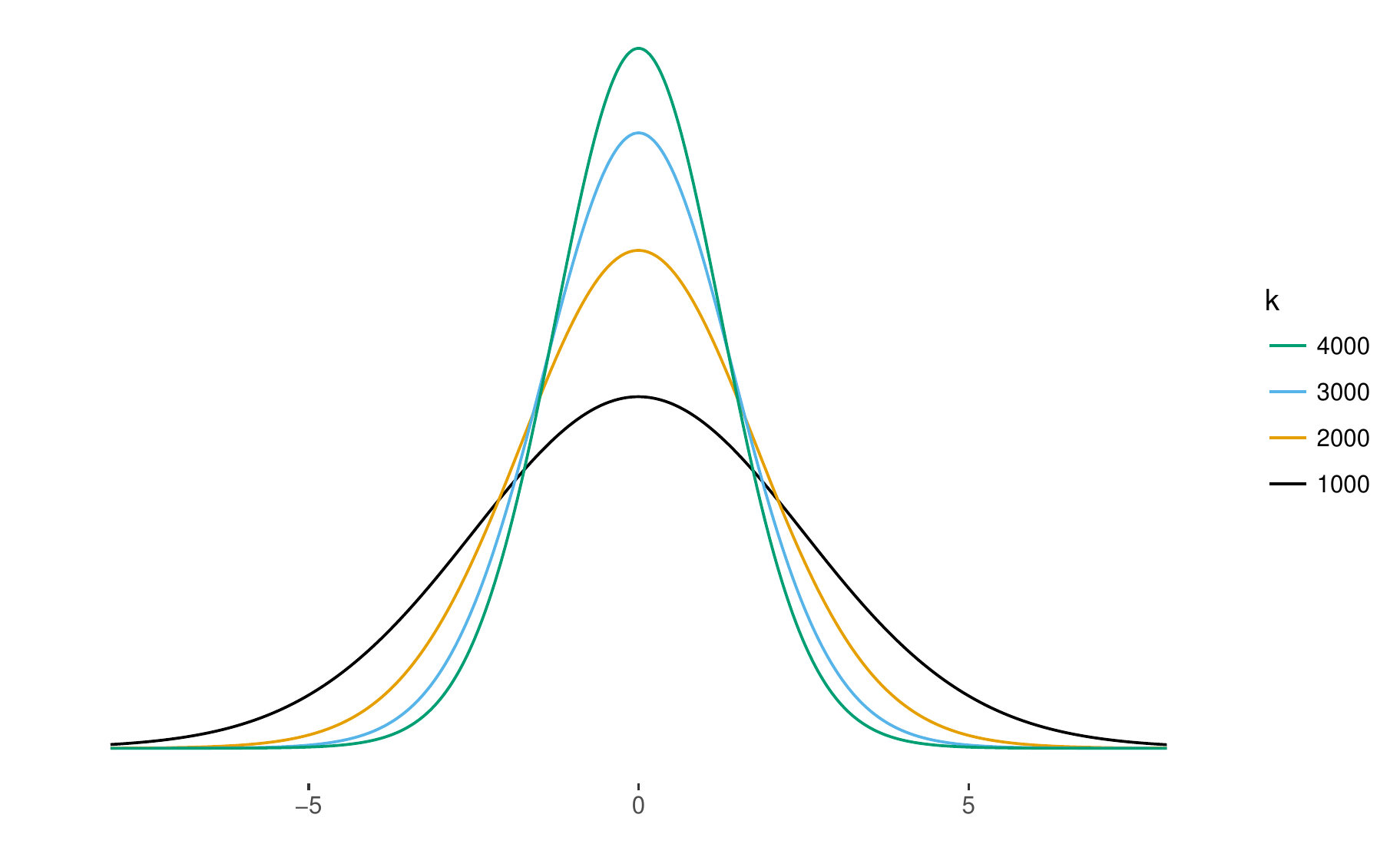}}
		\caption{Ternary random indices}
		\label{fig:ri_dist_ternary}
	\end{center}
\end{subfigure}
\hfill 
\begin{subfigure}[b]{1\columnwidth}
\centering
	\begin{center}
		\centerline{\includegraphics[width=\linewidth]{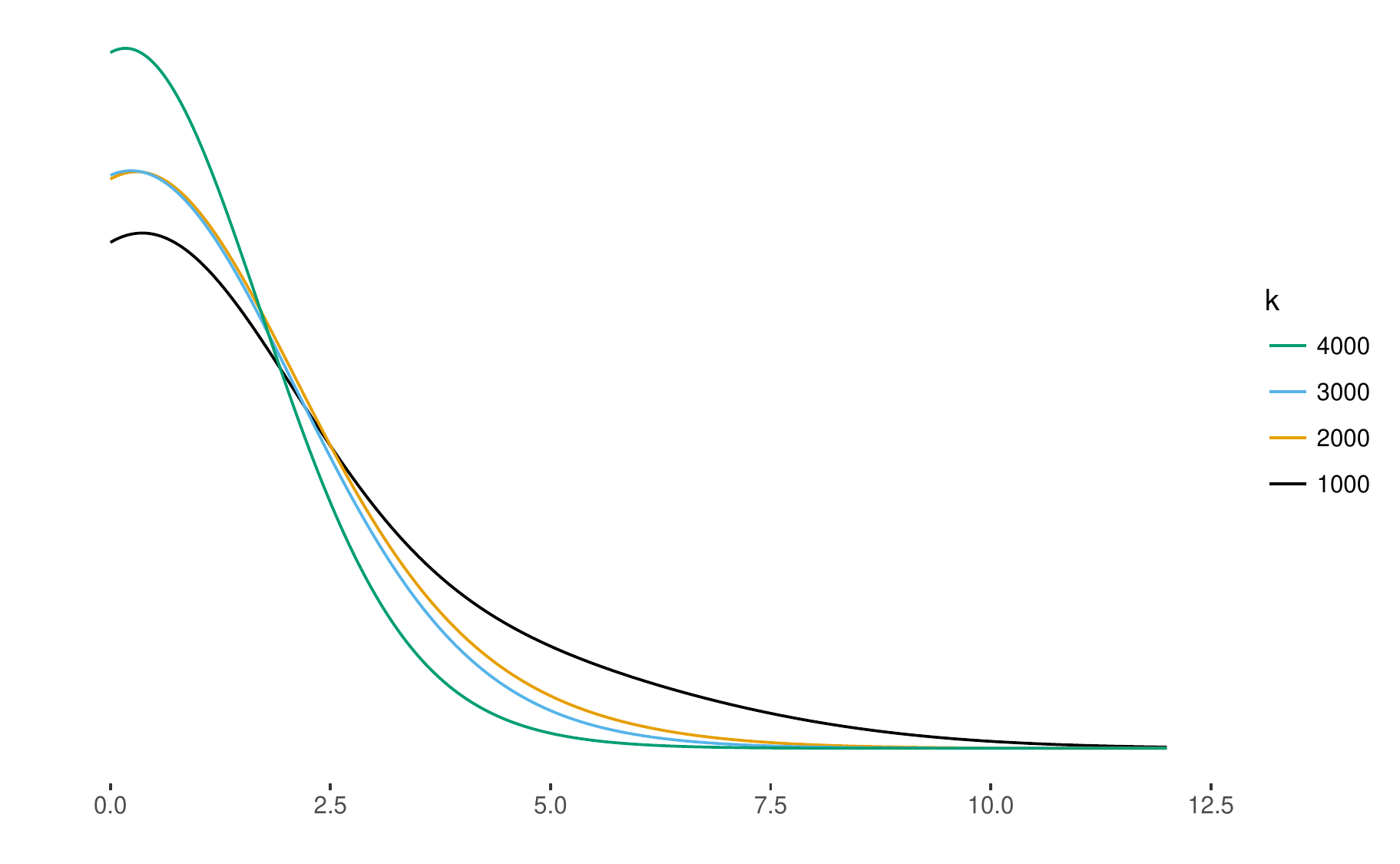}}
	    \caption{Binary random indices}
		\label{fig:ri_dist_binary}
	\end{center}

\end{subfigure}\hfill
\caption{Inner product distribution for different values of random index dimension $k$, aggregating the 
distributions for multiple numbers of non-zero entries. Ternary random indices \ref{fig:ri_dist_ternary} 
have entries with values $\in \{-1,0,1\}$, while binary random indices have entries with values $\in \{0,1\}$.}
\label{fig:ri_dist}
\end{figure*}

\noindent Having more non-zero entries adds to the proabability of having collisions 
(same columns between two vectors having non-zero values), which leads to larger inner products, 
nevertheless the inner product distribution does not deviate much from $0$. We will see later that 
having random mappings with more non-zero entries is not necessarily bad as it adds to the robustness 
of the input encoding -- making it easier for the neural network model to distinguish between different 
input patters despite the collisions.

If we take the distributions in figure \ref{fig:ri_1000} as a profile for inner product concentration for 
a random index with dimension $k$ (in this case 1000), and we vary this $k$. We can see how the inner 
product becomes more concentrated on $0$ for higher dimensional random indices (figure \ref{fig:ri_dist}). 
In section \ref{sec:results} we will show how different sizes of random indices and number of non-zero entries 
affect the predictive capabilities of our neural network language models. One of the goals is to figure 
out if neural networks can still make predictions based on a compressed signal of its input, and to figure out 
if adding more non-zero entries (increasing the probability of collisions) hurts performance. Another 
question we explore is the importance (or irrelevance) of using binary (figure \ref{fig:ri_dist_binary}) 
vs ternary (figure \ref{fig:ri_dist_ternary}) random indices.

\subsection{Baseline Model Architecture}
\begin{figure}[ht]
	\vskip 0.2in
	\begin{center}
		\centerline{\includegraphics[width=\columnwidth]{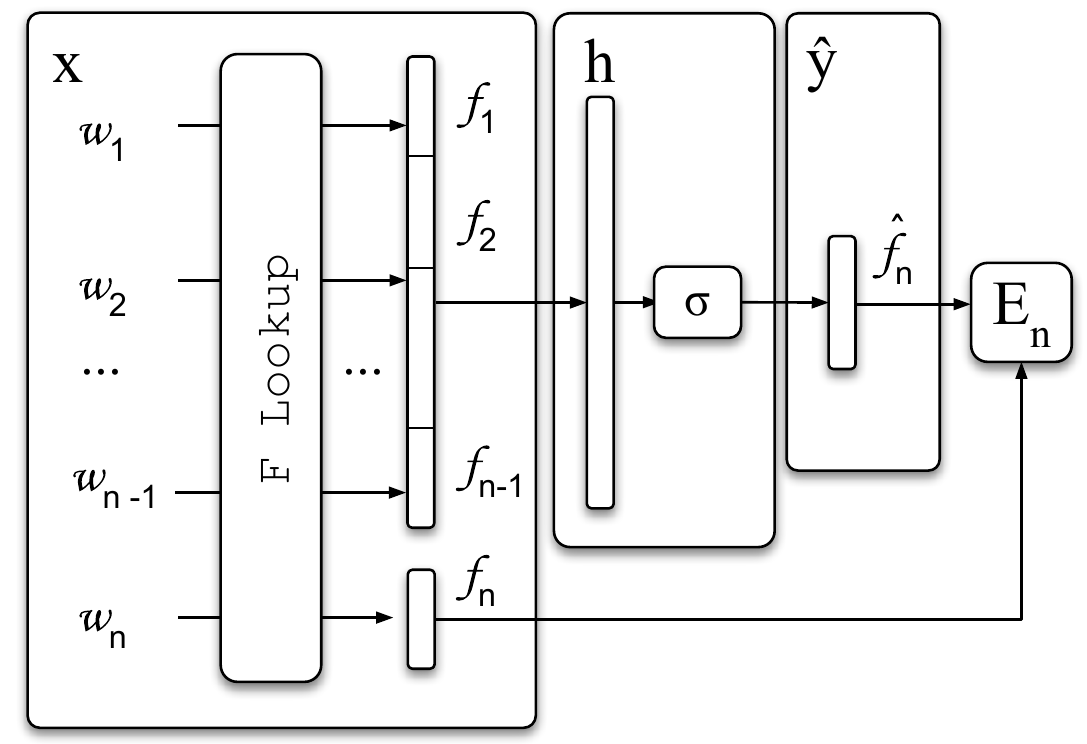}}
		\caption{Baseline feedforward energy model based on \cite{Bengio2003,Mnih2007}.}
		\label{fig:nnlm_energy}
	\end{center}
	\vskip -0.2in
\end{figure}

In order to assess the performance of our added random projection module, we build a \textbf{baseline} model
\textit{without} random projections. The architecture for this model is based in the one in \cite{Bengio2003} but
instead of predicting the probability of each word directly we use the energy-based definition from Mnih's work
\cite{Mnih2007}. The baseline model learns to predict the feature vector (embedding) for a target word along with a
probability distribution for all the known words conditioned on the learned feature
vectors of the input context words. Figure \ref{fig:nnlm_energy} shows an overview of our baseline model. 

The baseline language model in figure \ref{fig:nnlm_energy} converts each word into a vector of 
real-valued features using a feature lookup table $F$. The goal is to learn a set of feature vectors in 
$F$ that are good at predicting the feature vector for the target word $w_n$. Each word in a context of 
size $n-1$ is converted to $n-1$ $m\text{-dimensional}$ feature vectors that are then concatenated into 
a $(n-1) \times m$ vector and passed to the next layer $h$ that applies a non-linear transformation:

\begin{equation}
    h_j = \sigma\left(W_hf + b_h\right)
\end{equation}

where $f$ is a vector of $n-1$ concatenated feature vectors, $b$ is a bias for the layer $h$ and 
$\sigma$ is a non-linear transformation (e.g. sigmoid, hyperbolic tangent, or rectifier linear unit 
(ReLU)). We found ReLU units (equation \ref{eq:relu}) to perform well and at an increased training 
speed, so we use this transformation in our models. 

\begin{equation}
    f(x) = max(0,x)
    \label{eq:relu}
\end{equation}

The result of the non-linear transformation in $h$ is then passed to a linear layer $\hat{y}$ that 
predicts the feature vector $\hat{f}_n$ for the target word $w_n$ as follows:

\begin{equation}
    \hat{y}_n = W_{\hat{y}}h + b_{\hat{y}} 
\end{equation}

The energy function $E$ is defined as the dot product between the predicted feature vector 
$\hat{f}_n$, and the actual current feature vector for the target word $f_n$:

\begin{equation}
    E_n = \hat{f}^T_{n}f_n
    \label{eq:energy}
\end{equation}

To obtain a word probability distribution in the network output, the predicted feature vector is compared 
to the feature vectors of all the words in the known dictionary by computing the energy value 
(equation \ref{eq:energy}) for each word. This results in a set of similarity scores which are exponentiated 
and normalised to obtain the predicted distribution for the next word:

\begin{equation}
    P(w_n|w_1:n-1) = \frac{\exp(E_n)}{\sum_j^{|V|} \exp(E_j)}
    \label{eq:word_probability}
\end{equation}

The normalisation factor is the sum of the energy values for all the words in the dictionary. Contrary 
to the work in \cite{Mnih2009}, we didn't find any advantage in adding bias values based on the 
frequencies of the known words in the dictionary, so we use a pure energy-based definition.

The described baseline model can be though of as the standard feedforward one, as defined in \cite{Bengio2003} 
with an added linear layer $\hat{y}$ of the same size as the embedding dimensions $m$. Furthermore, 
instead of using two different sets of embeddings to encode context words and compute the word probability 
distributions, we learn a single set of feature vectors $F$ that are shared between input and output layers 
to compute the energy values and consequently the probability distribution over words.

\subsection{NRP Model Architecture}
We incorporate our random projection encoding into the previously defined energy-based neural network 
language model following the formulation in \cite{Mnih2007,Mnih2009}. We chose this formulation because 
it allows us to use maximum-likelihood training and evaluate our contribution in terms of \textit{perplexity}. 
This makes it easier to compare it to other approaches that use the same intrinsic measure of model performance. 
It is well known that computing the normalisation constant (the denominator in equation
\ref{eq:word_probability}) is expensive, but we are working with relatively simple models, 
so we compute this anyway. It should be noted that the random projection definition is compatible 
with much more economical approximation methods such as
\acrfull{nce} \cite{Mnih2012}, and as such, we plan to test this method in the future along with other evaluation
methods.

\begin{figure}[ht]
	\vskip 0.2in
	\begin{center}
		\centerline{\includegraphics[width=\columnwidth]{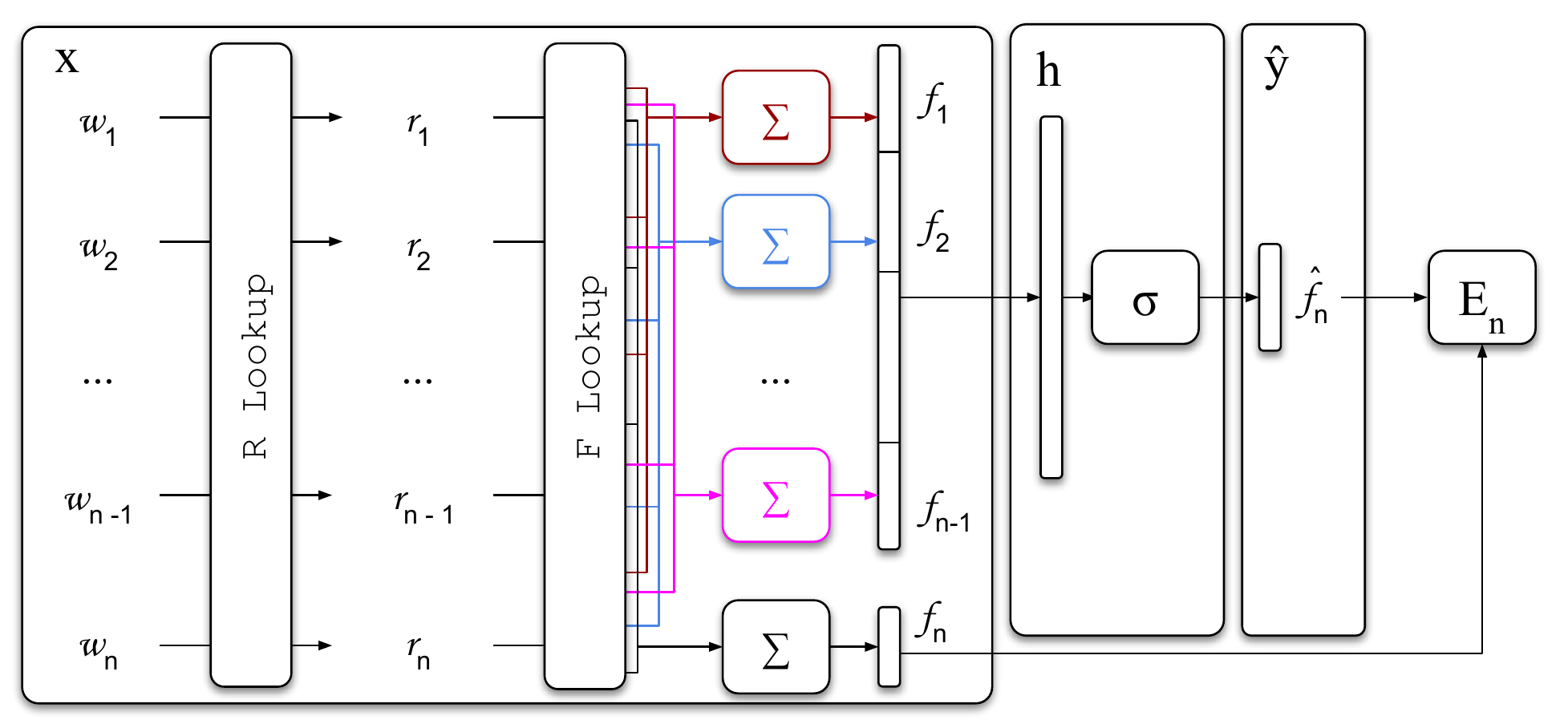}}
		\caption{Neural random projection feedforward energy model.}
		\label{fig:nrp}
	\end{center}
	\vskip -0.2in
\end{figure}

Having defined the baseline model, we now extend it with a random projection lookup layer. 
The overview of the resulting architecture can be seen in figure \ref{fig:nrp}.

There are two major differences between this and the previous model. First of all, we include a random index 
lookup table that generates a new random index for each new word that it encounters. Tho get the feature 
representation for a given word $w_i$ we multiply its $k-\text{dimensional}$ random index by the $k \times m$ 
feature matrix $F$. The resulting vector representation for a word $w_i$ is given by the sum of all the 
feature vectors extracted for each non-zero entry in the random index. Much like in existing models, only 
the rows corresponding to the non-zero entries are modified during training for each input sample. This 
adds to the computational complexity of the first layer since we are no longer selecting a single feature 
vector for each word, but it is still quite efficient ,since for $n-1$ words in a context, the feature 
vectors for the context can be retrieved as single sparse multiplication. 

The second difference is that to compute the output probability normalisation constant, we can no longer 
compute the dot product between the predicted feature vector $\hat{f}_n$ and the feature matrix $F$ as 
$\hat{f}_n^T F$ because we no longer have one per word. Instead, we must first find the feature vectors 
for all the words by multiplying all the random indexes for all the words stored in the random index 
lookup $R$ by $F$ as $F'=R^TR$ and then find the normalisation constant by computing the dot product 
between the predicted feature vector and $F'$. The probability normalisation function is thus given by:

\begin{equation}
    \sum_j^{|V|} \exp(\hat{f}_j^Tr_j),
\end{equation}

where $r_j$ (as we can see in figure \ref{fig:nrp}) is the sparse random index for word $w_j$. While all 
these operations can be implemented using sparse matrix multiplications, the computational complexity of 
each training or inference step scales linearly with the number of non-zero entries in the random indices $s$. 
As we stated previously, techniques like \acrshort{nce},can alleviate the problem since the issue lies in 
the fact that we need to compute the partition term to learn an output probability distribution. 

The fact that we can learn features for a vocabulary incrementally does not eliminate the fact that we 
need to compute the normalisation constant or the fact that the model does not deal with out-of-vocabulary 
words. Nevertheless, we are adding to neural network models, is the possibility of building predictive models based on a 
compression of the input space and enabling the possibility for the encoding of more complex 
(and possibly structured) inputs using random projections similarly to \cite{Basile2011}.

Since all models output probability distributions, they are trained using a maximum likelihood criterion:

\begin{equation}
\mathcal{L} = \frac{1}{T} \sum_{t} \log P(w_t,w_t-1,\ldots,w_t-n+1;\theta)
\end{equation}

Where $\theta$ is the set of parameters learned by the neural network in order to maximise a corpus likelihood. 

In the following section, we present our experimental setup. We start by describing the used dataset, evaluation metrics, along with a brief description of the hyper-parameters found in exploratory model fine-tuning. We proceed to show how random index encoding influences model performance and compare the model with the previously described baseline.

\section{Experimental Results}
\label{sec:results}

\subsection{Experimental Setup}
We evaluate our models using a subset of the \textit{Penn Treebank} corpus, more specifically, the portion containing
articles from the Wall Street Journal \cite{Marcus1993}. We use a commonly used set-up for this corpus, with pre-processing
as in \cite{Mikolov2011} or \cite{Kim2016}, resulting in a corpus with approximately 1 million tokens (words) and a
vocabulary of size  $|V| = 10\text{K}$ unique words The dataset is divided as follows: sections 0-20 were used as training 
data ($930$K tokens), sections 21-22 as validation data (74K tokens) and 23-24 as test data ($82$K tokens). All words
outside the 10K vocabulary were mapped to a single token \textit{$<unk>$}. We train the models by dividing the PTB corpus 
into \textit{n-gram} samples (sliding windows) of size $n$. All the models here reported were $5-text{g-ram}$ models. To 
speed up training and evaluation, we batch the samples in mini-batches of size $128$ and the training set is randomised 
\textit{prior} to batching. 

We train the models by using the \textit{cross entropy} loss function. \textit{Cross entropy} measures the 
efficiency of the optimal encoding for the approximating (model predicted) distribution $q$ under the true (data)
distribution $p$:

\begin{equation}
	H(p,q) = -\sum_x p(x)log(q(x))
\end{equation}

We evaluate and compare the models based on their perplexity:

\begin{equation}
    \text{PPL} = \exp \left( -\frac{1}{N} \sum_{w_{1:n}} log P(w_n|w_{1:n-1}) \right) 
\end{equation}

where the sum is performed over all the \textit{n-gram windows} of size $n$. Another way to formulate perplexity is as follows: 

\begin{equation}
    2^{H(p,q)} = 2^{-\sum_x p(x)log(q(x))} \footnote{If $\ln$ is used instead of $\log$ to compute the cross-entropy, perplexity must be computed as $e^{H(p,q)}$ instead.}
\end{equation}

Perplexity is essentially a geometric average of inverse probabilities; it measures the predictability of some 
distributions, given the number of possible outcomes of an equally (un)predictable distribution, if each of those 
outcomes was given the same probability. (Example: if a distribution has perplexity $6$, it is as unpredictable as 
a fair dice. Since all the models are trained using mini-batches, all perplexities are model perplexities, 
this is, the average perplexity for all the batches.)

Along with model performance, we also report the approximate number of trainable parameters of each model 
configuration which can be computed as:

$$\#p =(|x| \times m) + ((m \times (n-1)) \times h + h) + (h \times m + m)$$

where $|x|$ s the dimension of input vectors, $m$ is the dimension of the feature lookup layer (or embeddings), 
$n$ is the \textit{n-gram} size (in this case $5$), and $h$ is the number of hidden units. 

We train our models using \acrfull{sgd} without momentum. In early experiments, we found that \acrshort{sgd} 
produced overall better models and generalised better than adaptive optimisation procedures such as \textit{ADAM}
\cite{Kingma2014}. Model selection and early stopping is based on the perplexity measured on the validation set.

We used a step-wise learning rate annealing ,where the \textit{learning rate} is kept fixed during a single epoch, 
but we shrink it by $0.5$ every time the validation perplexity increases at the end of a training epoch. If the 
validation perplexity does not improve, we keep repeating the process until a given number of epochs have 
passed without improvement (early stop). The \textit{patience} parameter (number of epochs without improvement 
before stopping) is set to $3$ and it resets if we find a lower validation perplexity at the end of an epoch. 
We consider that the model \textbf{converges} when it stops improving and the \textit{patience} parameter reaches $0$

As for \textit{regularisation}, \textit{dropout} \cite{Srivastava2014} and $\ell^2$ weight penalty performed similarly. 
We opted for using \textbf{dropout} regularisation in all the models. Dropout is applied to all weights including the 
feature lookup (embeddings). In early experiments, we used a small dropout probability of $0.05$ following the 
methodology in \cite{Pham2016} (where the authors use smaller dropout probabilities due to the Penn Treebank corpus 
being smaller. We later found that,dropout probabilities interact non-linearly with other architectural parameters 
such as embedding size and number of hidden units. Higher dropout probability values can be used (and give us better 
models) in conjunction with larger embedding and hidden layer sizes.

\begin{figure}[ht]
	\vskip 0.2in
	\begin{center}
		\centerline{\includegraphics[width=\columnwidth]{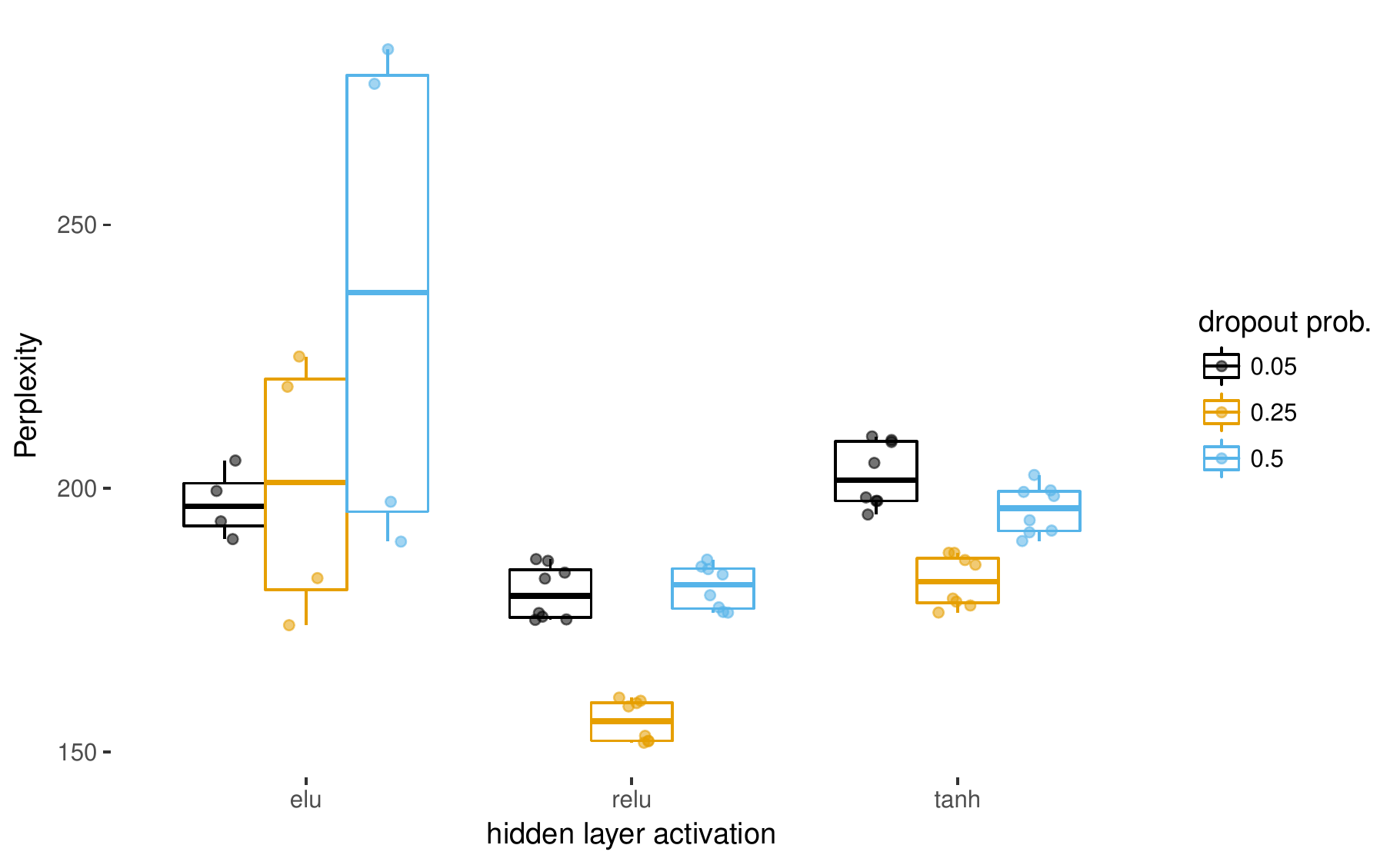}}
		\caption{Average \textit{test} perplexity of best NRP models for different hidden activation functions for a feature lookup size (embeddings) $m=128$, a number of hidden units $h=256$, random indices of dimension $k=5000$ and number of non-zero units $s=8$}
		\label{fig:nrp_actfn}
	\end{center}
	\vskip -0.2in
\end{figure}

We found that Rectifying Linear Units (ReLU) activation units \cite{He2015} performed better than saturating 
non-linear functions like \textit{hyperbolic tangent}, or alternatives like Exponential Linear Units (ELUs)
\cite{Clevert2015}. Figure \ref{fig:nrp_actfn} shows the distribution of test perplexity for models using random 
projections with different activation functions on the non-linear projection layer. Overall, ELUs are the most 
sensitive to larger dropout regularisation and didn't produce significantly better results than the models 
using the \textit{hyperbolic tangent}. Rectifier linear units seem to outperform other types of non-linearities 
both in our baseline and the models using random projections (NRP).

One problem encountered when using ReLU units was that they caused gradient values to \textit{explode} easily during
training. To mitigate this, we added \textit{local} gradient norm clipping: each time a gradient norm is greater 
than a given threshold (in our case $1.0$) we clip the gradient to the threshold value. Clipping gradients by 
global norm resulted in worse models, so we clip their value by their individual norms. Gradient clipping 
allowed us to start with higher \textit{learning rate} values. Setting the initial \textit{learning rate} to 
$0.5$ yielded the best results, both in terms of model validation perplexity and number of epochs for convergence. 

All network weights are initialised randomly on a range $[-0.01,0.01]$ in a random uniform fashion. Bias values are
initialised with $0$. We found it beneficial to initialise the non-linear layers with the procedure described 
in \cite{He2015} for the layers using \textit{ReLU} units.
 
We performed multiple experiments using different sizes for both feature vectors (embeddings) and hidden layer size, 
but larger feature vector spaces do not lead to better models unless more aggressive regularisation is applied 
(higher dropout probability).

\subsection{NRP properties: initial experiments}
Upon selecting a good set of hyper-parameters, we initially explored the effects of different random index (RI)
configurations in the model. This influence is measured in terms of perplexity (less is better). In these first 
experiments, the embedding size is set to $m=128$ and the hidden layer to $h=256$, while the dropout probability is also fixed
to $0.05$. We span three different parameters, varying the embedding size $m \in \{ 128,256,512,1024 \} $, the RI dimension
$k \in [1000,10000]$ and the RI number of non-zero entries $s \in \{2,8,16\} $. 
We run each model to convergence 
(early stop using validation perplexity). We save the best models upon convergence based on validation perplexity 
and present the perplexity scores on the test set. 

\begin{figure}[ht]
	\vskip 0.2in
	\begin{center}
		\centerline{\includegraphics[width=\columnwidth]{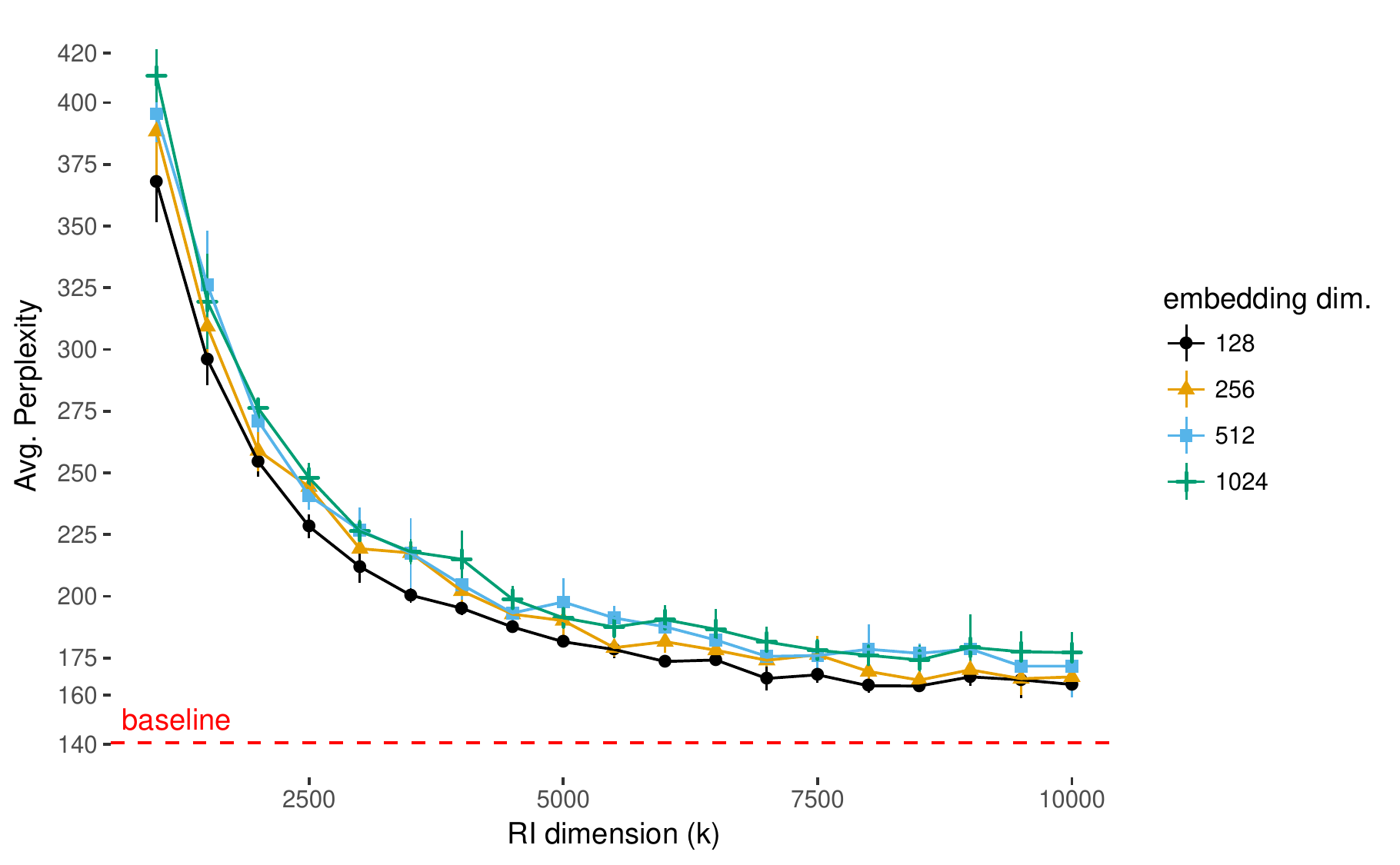}}
		\caption{Average \textit{test} perplexity of best models selected based on 
		\textit{validation} perplexity for multiple embedding sizes $m \in \{128,256,512,1024\}$ 
		aggregated by number of non-zero entries $ s \in \{2,8,16\} $. The number of hidden units was set to $256$.}
		\label{fig:k_curve_m}
	\end{center}
	\vskip -0.2in
\end{figure}

Figure \ref{fig:k_curve_m} shows the results for the average test perplexity scores aggregated by the number 
of non-zero entries in the RIs. The perplexity score follows an exponential decay as the RI dimension $k$ increases. 
From $k>=5000$, the perplexity values converge to an approximate range $PPL \approx [170,180]$ (see table 
\ref{tab:k_curve_m}). This is worse than the best baseline model, but we subsequently found out that a more aggressive 
regularisation process improves both the NRP and the baseline models. 

Increasing the size of the embedding features $m$ seems to yield worse perplexity. We found that 
models with more parameters where overfitting the training set, thus explaining the worse results. 
More aggressive regularisation (higher dropout probabilities) seems to alleviate the problem, but for 
embedding and hidden layer sizes of $1024$ there is either no improvement, or the improvement is marginal. 

\begin{table}[!htbp] \centering 
  \caption{Average test perplexity for the best models selected based on validation perplexity 
  aggregated by number of active units (see figure \ref{fig:k_curve_m}). The number of hidden units was set to $256$.}
  \label{tab:k_curve_m} 
  \vspace{1em}
  \centering
\begin{tabular}{@{\extracolsep{1pt}} l@{\hspace{0.9\tabcolsep}}l@{\hspace{0.9\tabcolsep}}lr@{\hspace{0.9\tabcolsep}}rr@{\hspace{0.9\tabcolsep}}r} 
& &  &\multicolumn{2}{l}{PPL} & \multicolumn{2}{l}{Epoch}   \\
$k$ & $m$ & $\#p$ & Avg. & SD & Avg. & SD \\ 
\hline \\[-1.8ex] 
$5,000$  & $128$   & $0.8$  & $182$ & $0.2$ & $12$ & $2.0$ \\ 
         & $256$   & $1.6$  & $190$ & $5.9$ & $9$  & $1.0$ \\ 
         & $512$   & $3.2$  & $198$ & $9.6$ & $8$  & $0.0$ \\ 
         & $1,024$ & $6.3$  & $191$ & $2.9$ & $8$  & $0.0$ \\ 
$7,500$  & $128$   & $1.1$  & $168$ & $3.2$ & $12$ & $0.5$ \\ 
         & $256$   & $1.6$  & $176$ & $7.6$ & $8$  & $0.0$ \\ 
         & $512$   & $3.2$  & $176$ & $5.2$ & $7$  & $0.5$ \\ 
         & $1,024$ & $8.9$  & $178$ & $4.6$ & $7$  & $1.2$ \\ 
$10,000$ & $128$   & $1.4$  & $164$ & $1.4$ & $13$ & $2.5$ \\ 
         & $256$   & $2.8$  & $167$ & $8.1$ & $8$  & $0.0$ \\ 
         & $512$   & $5.7$  & $172$ & $12.6$& $8$  & $0.6$ \\ 
         & $1,024$ &  $11$  & $177$ & $8.4$ & $7$  & $1.0$ \\ 
\addlinespace
baseline & & & & \\
\hline \\[-1.8ex] 
$10,000$ & $128$ & $1.4$ & $141$ & $1.4$   & $14$ & $1.1$  \\ 
\end{tabular} 
\end{table} 

\begin{figure}[ht]
	\vskip 0.2in
	\begin{center}
		\centerline{\includegraphics[width=\columnwidth]{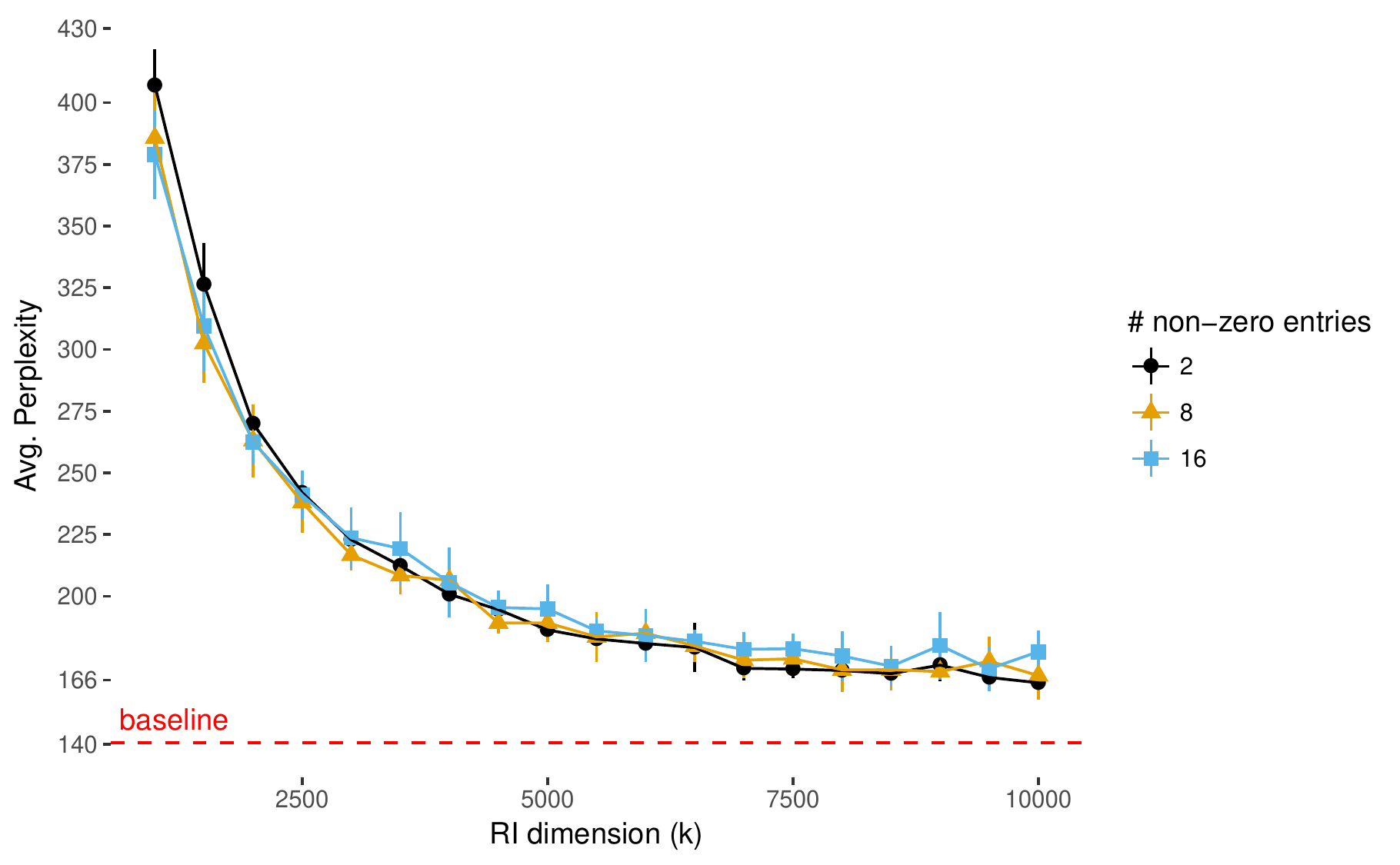}}
		\caption{Average \textit{test} perplexity of the best models selected based on 
		\textit{validation} perplexity for random indices with a number of non-zero entries 
		$ s \in \{2,8,16\} $ aggregated by embedding size $m \in \{128,256,512,1024\}$. The number of 
		hidden units was set to $256$.}
		\label{fig:k_curve_s}
	\end{center}
	\vskip -0.2in
\end{figure}

We then looked at the influence of number of non-zero entries $s$ in the random index vectors for the NRP models, 
by aggregating the results by embedding size. Figure \ref{fig:k_curve_s} shows that different values of $s$ 
do not affect the performance significantly. We should note that this parameter does affect the model performance 
in terms of perplexity (although marginally) and should be optimised if the goal is to produce the best possible model  
for a given data set. 

The number of epochs to convergence is correlated with the number of trainable parameters 
(as we will see when comparing these early explorations with later results), but it does not differ 
significantly between NRP and baseline models (see tables \ref{tab:k_curve_m} and \ref{tab:k_kurve_s}).


\begin{table}[!htbp] \centering 
  \caption{Average \textit{test} perplexity of the best models selected based on validation perplexity, 
  aggregated by embedding size $m$ (see figure \ref{fig:k_curve_s}) for multiple values of input vector 
  dimension $k$ and number of non-zero input values $s$. The number of hidden units was set to $256$. 
  Dropout probability is set to $0.05$. $\#p$ is the approximate number of trainable parameters in millions.} 
  \label{tab:k_kurve_s} 
  \centering
  \vspace{1em}
\begin{tabular}{@{\extracolsep{1pt}} l@{\hspace{0.9\tabcolsep}}l@{\hspace{0.9\tabcolsep}}lr@{\hspace{0.9\tabcolsep}}rr@{\hspace{0.9\tabcolsep}}r} 
& & &\multicolumn{2}{l}{PPL} & \multicolumn{2}{l}{Epoch}   \\
$k$      & $s$  & $\#p$ & Avg. & SD & Avg. & SD \\ 
\hline \\[-1.8ex] 
$5,000$  & $2$  & $0.8$ &  $186$ & $3.4$  & $10$ & $3.3$  \\ 
         & $8$  &       &  $189$ & $7.8$  & $9$  & $1.9$  \\ 
         & $16$ &       &  $195$ & $10.0$ & $9$  & $1.5$  \\ 
$7,500$  & $2$  & $1.1$ &  $170$ & $3.8$  & $9$  & $2.5$  \\ 
         & $8$  &       &  $175$ & $6.2$  & $9$  & $3.1$  \\ 
         & $16$ &       &  $179$ & $6.0$  & $9$  & $2.5$  \\ 
$10,000$ & $2$  & $1.4$ &  $165$ & $1.8$  & $9$  & $3.5$  \\ 
         & $8$  &       &  $168$ & $9.8$  & $8$  & $1.2$  \\ 
         & $16$ &       &  $177$ & $8.7$  & $9$  & $3.1$  \\ 
\addlinespace
baseline & & & &\\
\hline \\[-1.8ex] 
$10,000$ & $1$  & $1.4$ & $141$ & $1.4$   & $14$ & $1.1$  \\ 
\end{tabular} 
\end{table}

\begin{figure}[ht]
	\vskip 0.2in
	\begin{center}
		\centerline{\includegraphics[width=\columnwidth]{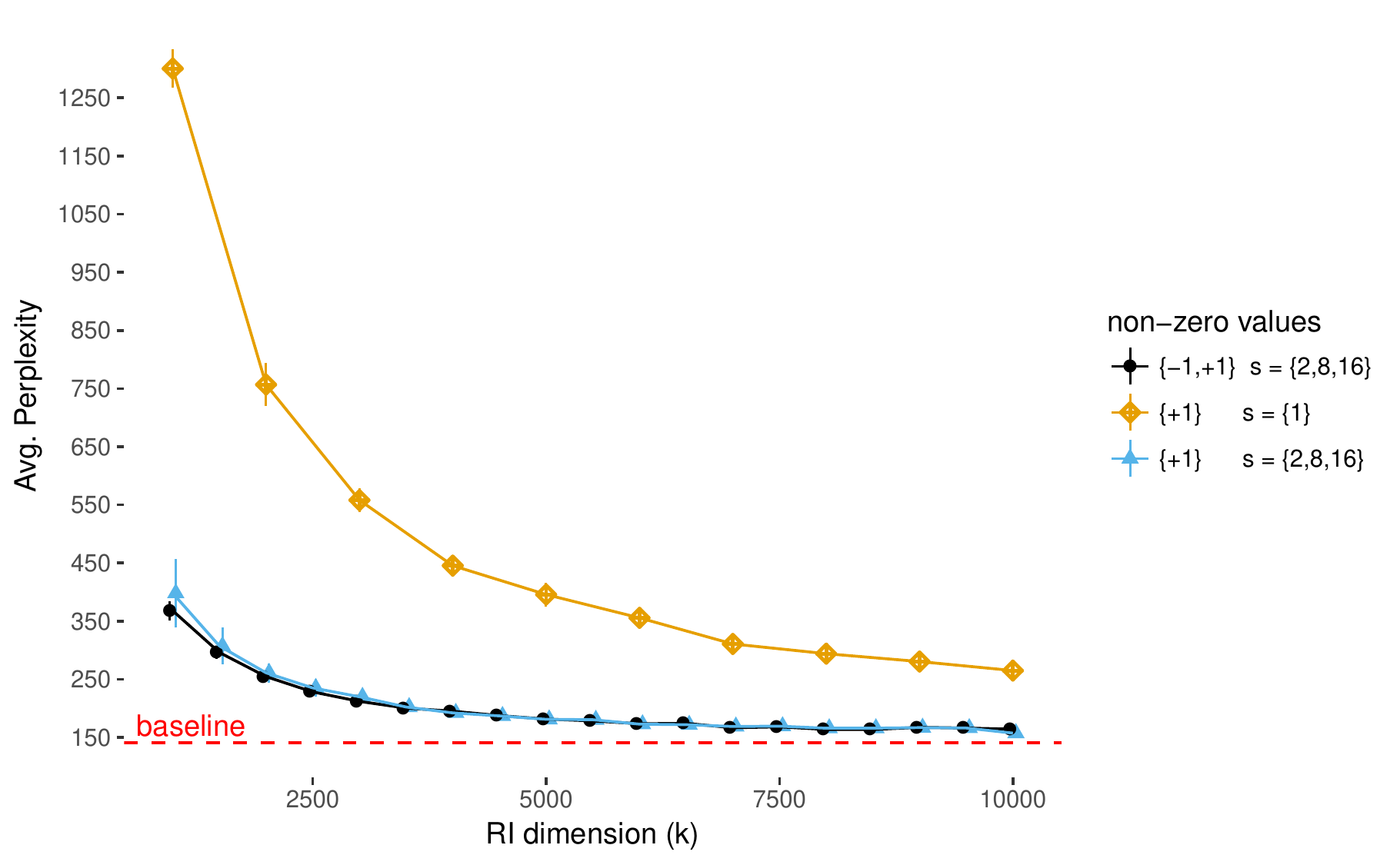}}
		\caption{Average \textit{test} perplexity of best models selected based on \textit{validation} perplexity for random indices constructed from typical ternary random indices with non-zero entries having values $\in \{+1,1\} $; random indices with a single positive non-zero entry $\{+1\} s \in \{1\}$; and binary random index vectors with non-zero entries $\in \{+1\} $. }
		\label{fig:k_curve_symmetry}
	\end{center}
	\vskip -0.2in
\end{figure}

Finally, in our early explorations, we looked at two different questions. First, we wanted to 
find out if there was a difference in performance between NRP models using ternary random indices 
with values $\in \{-1,0,1\}$) and models using binary random indices (with values 
$\in \{0,1\}$). Second, we wanted to find if redundancy (having more non-zero entries) is key to NRP model 
performance. To answer this second question, we tested models that used random indices with a single non-zero value. 
Under this scenario, different words can share the same feature vector. Figure \ref{fig:k_curve_symmetry} 
shows the results for this experiment, in which we tested models with these $3$ configurations.

We can see that models having a single non-zero positive entry $s=1$ do much worse than any other configuration. 
Having just two non-zero entries $s>=2$ seems to be enough to improve the model perplexity. A second result we got 
from this experiment was that using binary and ternary random indices yields similar results, with binary indices 
displaying an increased variance for lower values of random index dimension $k$. We think this increase in variance 
is due to the fact that lower-dimensional binary random indices have a higher probability of collisions and thus
disambiguation of such collisions becomes harder for our neural network models. As for why the results are similar, 
we should note that the near-orthogonality properties of random indices are still maintained whether or not we use 
ternary representations. Moreover, the feature layer (or embeddings) is itself symmetric, having equally distributed 
positive and negative weights, so the symmetry in the input space should not have much impact on the model performance.

\subsection{The role of regularisation}
\begin{figure}[ht]
	\vskip 0.2in
	\begin{center}
		\centerline{\includegraphics[width=\columnwidth]{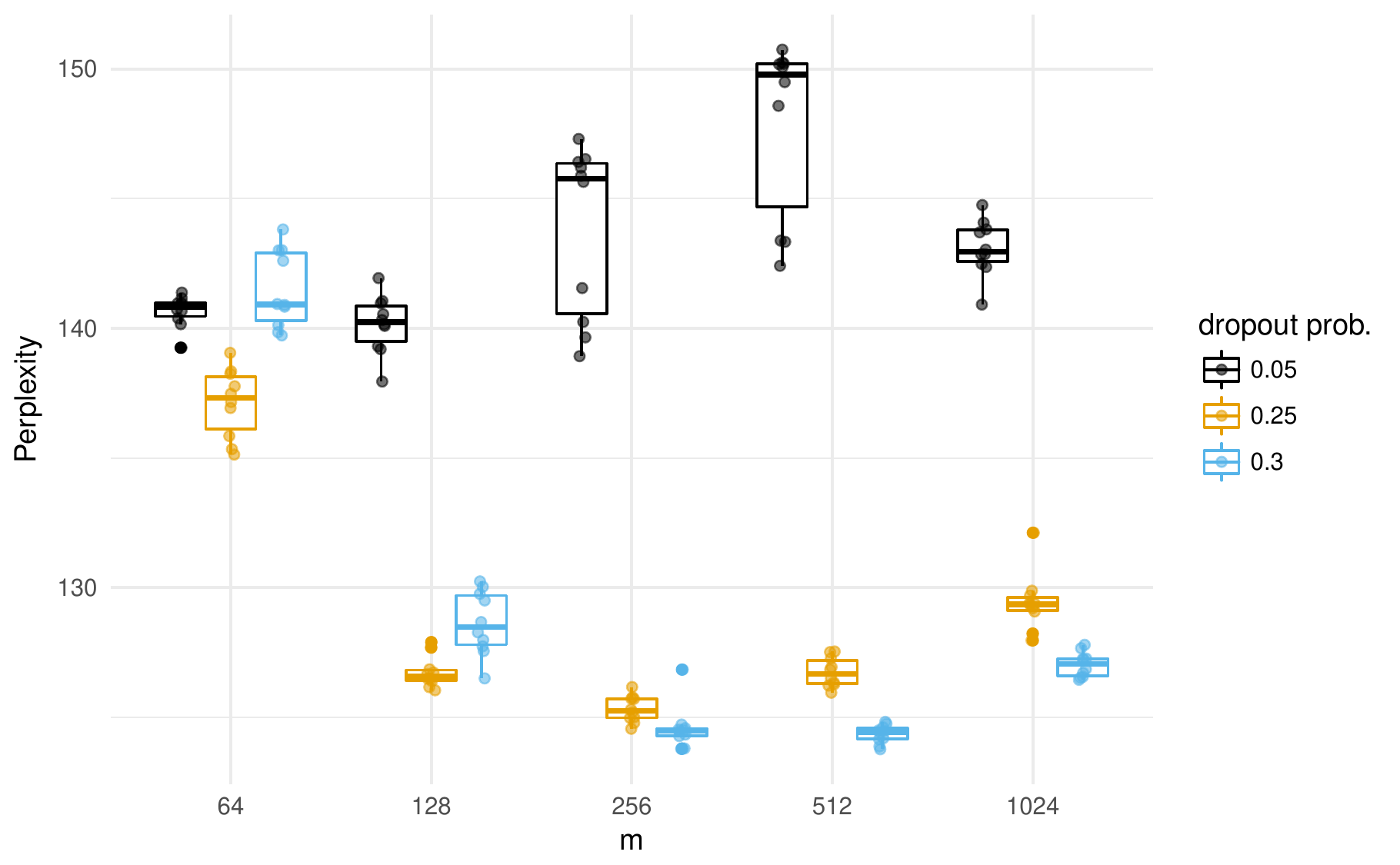}}
		\caption{Baseline model \textit{test} perplexity of best models selected based on \textit{validation} 
		perplexity for multiple values of embedding size $m$ and \textit{dropout probability}.}
		\label{fig:baseline_drop}
	\end{center}
	\vskip -0.2in
\end{figure}

In the next set of experiments, we look at the role of \textit{dropout probability} in both the baseline and 
NRP models. Figure \ref{fig:baseline_drop} shows the test perplexity scores for the baseline model using different 
embedding sizes $m$ and dropout probability values $p=\{0.05,0.25,0.3\}$. The number of hidden units is fixed at $h=256$. 

\begin{figure}[ht]
	\vskip 0.2in
	\begin{center}
		\centerline{\includegraphics[width=\columnwidth]{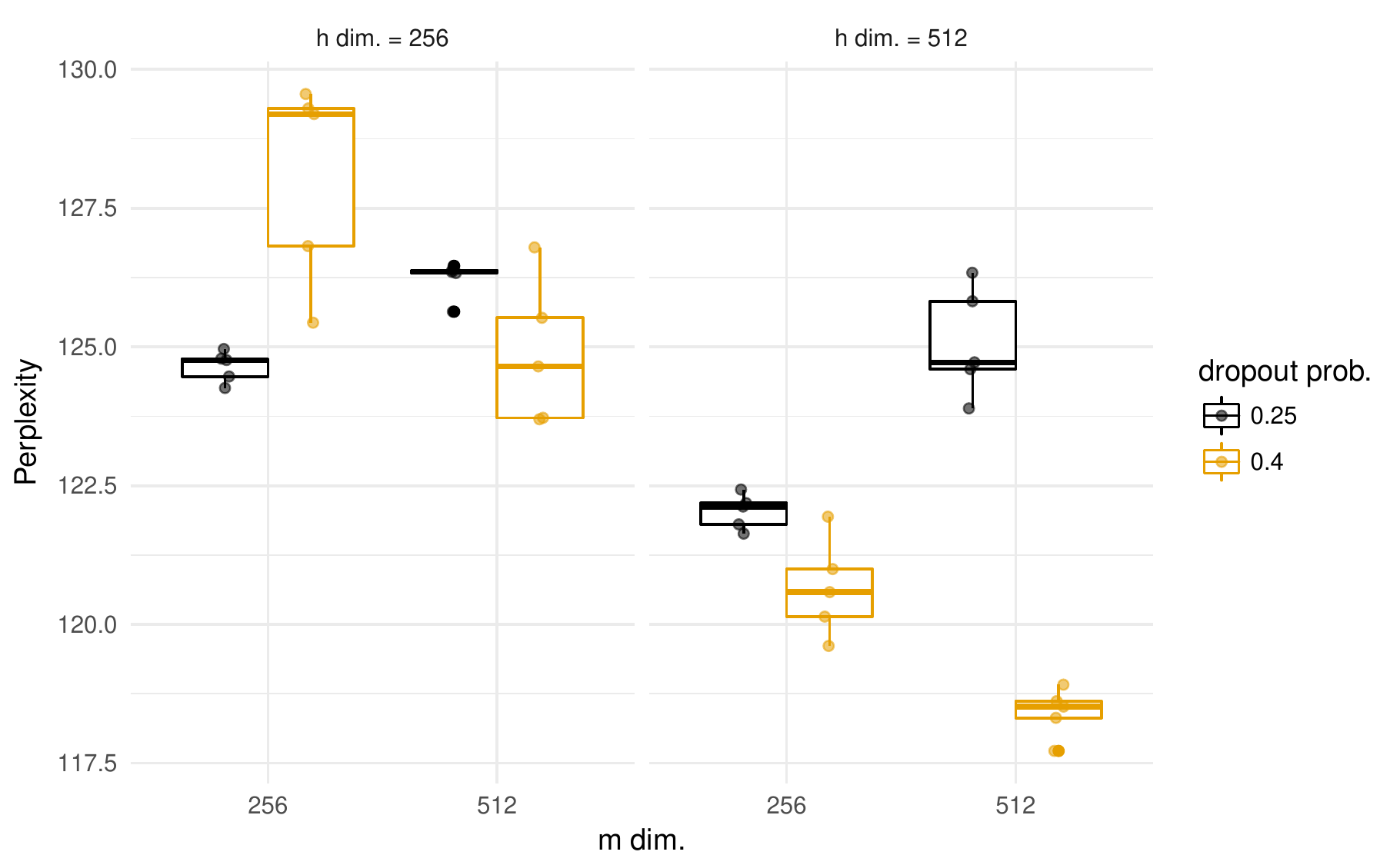}}
		\caption{Baseline model \textit{test} perplexity of best models selected based on 
		\textit{validation} perplexity for multiple values of embedding size $m$, hidden layer number 
		of units $h$ and \textit{dropout probability}.}
		\label{fig:baseline_best}
	\end{center}
	\vskip -0.2in
\end{figure}

We can see that keeping the \textit{dropout probability} at the previously used value $0.05$ makes the model 
perform worse for embedding size values $m>128$. We found out that larger models were heavily overfitting  
the training set, which explains why more aggressive dropout rates give us better results for larger feature spaces. 
After extensive testing of this hypothesis and parameter tuning, we recommend the use of higher dropout when increasing the size of the hidden layer (figure \ref{fig:baseline_best}, and table \ref{tab:best_baseline}). 

\begin{figure}[ht]
	\vskip 0.2in
	\begin{center}
		\centerline{\includegraphics[width=\columnwidth]{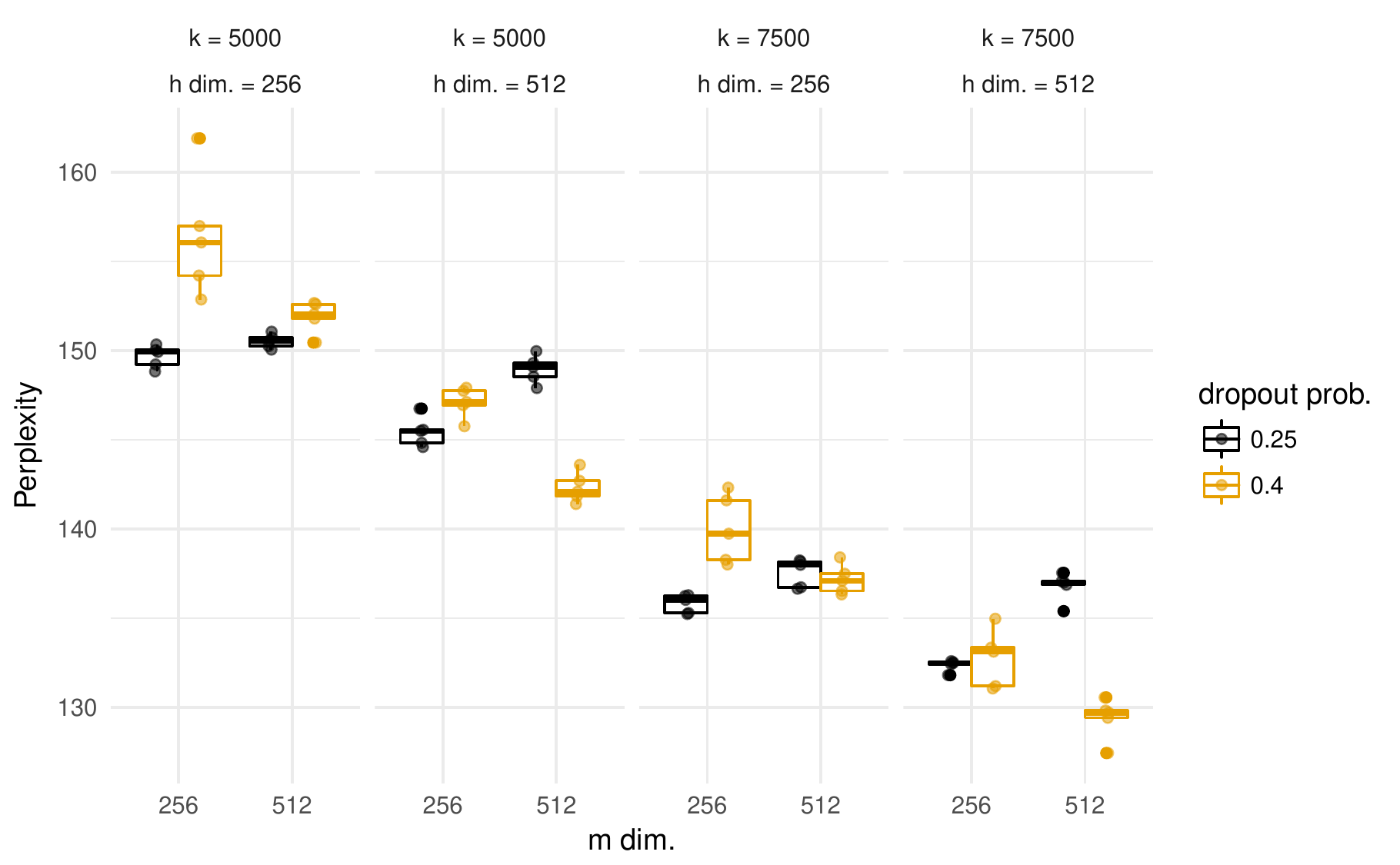}}
		\caption{NRP model \textit{test} perplexity of best models selected based on 
		\textit{validation} perplexity for multiple values of random index size $k$, 
		embedding size $m$, hidden layer number of units $h$ and \textit{dropout probability}.}
		\label{fig:nrp_best_all}
	\end{center}
	\vskip -0.2in
\end{figure}

Looking at the results for NRP models (figure \ref{fig:nrp_best_all}, table \ref{tab:best_nrp}), we can see that 
the best results are achieved with larger models and higher dropout probability. We should mention that even with 
half the embedding parameters (using random indices of dimension $k=5k$ to encode a vocabulary of $10k$ words), 
we get reasonably good models with test perplexities of $145$ to $150$ using only, $1.6$ to $1.9$ million
parameters (table  \ref{tab:best_nrp}). We can compare this with the perplexity scores obtain in \cite{Pham2016}
when using the same dataset and partitions. The model from \cite{Bengio2003} with approximately $4.5$ million
parameters gets a test perplexity of $147$ to $152$. 
This is of course not a fair comparison, because we would need to optimise both models using the best possible
hyper-parameter values and regularisation methods, but it does confirm our main hypothesis: that we can we can 
learn language models using a random projection of the input space.

\begin{table}[!htbp] \centering 
  \caption{Average test perplexity and convergence epoch for different configurations of the baseline 
  model with multiple values of embedding size $m$, number of hidden units $h$ and weight dropout 
  probability \textit{drop}. $\#p$ is the approximate number of trainable parameters in millions.} 
  \label{tab:best_baseline} 
\vspace{1em}
\begin{tabular}{
@{\extracolsep{1.5pt}} l 
@{\hspace{0.9\tabcolsep}} l 
@{\hspace{0.9\tabcolsep}} l 
@{\hspace{0.9\tabcolsep}} lr
@{\hspace{0.9\tabcolsep}}rr
@{\hspace{0.9\tabcolsep}} r} 
& & & & \multicolumn{2}{l}{PPL} & \multicolumn{2}{l}{Epoch}   \\
$m$    & $h$   & $\#p$ & \textit{drop} &Avg. & SD & Avg. & SD \\ 
\hline \\[-1.8ex] 
 $256$ & $256$ & $2.8$ & $.25$ & $125$ & $0.3$ & $22$ & $1.5$ \\ 
       &       &       & $.40$ & $128$ & $1.8$ & $22$ & $2.5$ \\ 
       & $512$ & $3.2$ & $.25$ & $122$ & $0.3$ & $19$ & $2.0$ \\ 
       &       &       & $.40$ & $121$ & $0.9$ & $23$ & $1.6$ \\ 
 $512$ & $256$ & $5.8$ & $.25$ & $126$ & $0.3$ & $20$ & $1.6$ \\ 
       &       &       & $.40$ & $125$ & $1.3$ & $22$ & $1.9$ \\ 
       & $512$ & $6.4$ & $.25$ & $125$ & $1.0$ & $16$ & $1.8$ \\ 
       &       &       & $.40$ & $\mathbf{118}$ & $0.4$ & $22$ & $1.3$ \\ 
\end{tabular} 
\end{table}

\begin{table}[!htbp] \centering 
  \caption{Average test perplexity and convergence epoch for different configurations of NRP models 
  with multiple values of \textit{random index} dimension $k$ embedding size $m$, number of hidden units $h$ 
  and weight dropout probability \textit{drop}. The number of non-zero entries in the RI is set fo $4$. $\#p$ 
  is the approximate number of trainable parameters in millions.} 
  \label{tab:best_nrp} 
  \vspace{1em}

\begin{tabular}{@{\extracolsep{1.5pt}} l @{\hspace{0.9\tabcolsep}} l @{\hspace{0.9\tabcolsep}} l 
@{\hspace{0.9\tabcolsep}} l @{\hspace{0.9\tabcolsep}} lr@{\hspace{0.9\tabcolsep}}rr@{\hspace{0.9\tabcolsep}} r} 
 & & & & & \multicolumn{2}{l}{PPL} & \multicolumn{2}{l}{Epoch}  \\
 $k$     & $m$    & $h$   & $\#p$ & \textit{drop} &Avg. & SD & Avg. & SD \\ 
\hline \\[-1.8ex] 
 $5,000$  & $256$ & $256$ & $1.6$ & $.25$ & $149$ & $0.6$ & $23$ & $1.2$ \\ 
          &       &       &       & $.40$ & $156$ & $3.5$ & $25$ & $3.4$ \\ 
          &       & $512$ & $1.9$ & $.25$ & $145$ & $0.8$ & $22$ & $1.7$ \\ 
          &       &       &       & $.40$ & $147$ & $0.9$ & $25$ & $1.1$ \\ 
          & $512$ & $256$ & $3.2$ & $.25$ & $150$ & $0.4$ & $21$ & $1.5$ \\ 
          &       &       &       & $.40$ & $151$ & $0.9$ & $24$ & $0.9$ \\ 
          &       & $512$ & $3.8$ & $.25$ & $149$ & $0.8$ & $19$ & $2.3$ \\ 
          &       &       &       & $.40$ & $\mathbf{142}$ & $0.9$ & $24$ & $1.0$ \\ 
\addlinespace
 $7,500$  & $256$ & $256$ & $2.2$ & $.25$ & $136$ & $0.5$ & $22$ & $1.3$ \\ 
          &       &       &       & $.40$ & $140$ & $1.9$ & $26$ & $2.9$ \\ 
          &       & $512$ & $2.5$ & $.25$ & $132$ & $0.3$ & $21$ & $0.8$ \\ 
          &       &       &       & $.40$ & $133$ & $1.6$ & $24$ & $2.3$ \\ 
          & $512$ & $256$ & $4.4$ & $.25$ & $137$ & $0.8$ & $21$ & $0.8$ \\ 
          &       &       &       & $.40$ & $137$ & $0.8$ & $24$ & $1.7$ \\ 
          &       & $512$ & $5.1$ & $.25$ & $137$ & $0.8$ & $18$ & $1.8$ \\ 
          &       &       &       & $.40$ & $\mathbf{129}$ & $1.2$ & $23$ & $1.5$ \\ 
\end{tabular} 
\end{table} 

Additional tests showed us that we cannot go much further with dropout probabilities, so we stuck to a 
maximum value of $0.4$, since it yielded the best results for both the baseline and models using random
projections. One weakness in our methodology was that hyper-parameter exploration was done based on grid search 
and intuition about model behaviour -- mostly because each model run is very expensive. While our goal is not to
find the best possible baseline and NRP models, but to show that we can achieve qualitatively similar results with
random projections and a reduced embedding space, in future work this methodological aspect will be improved by
replacing grid search by a more appropriate method such as \textit{Bayesian Optimisation} \cite{Snoek2012}. This
way, we can compare the best possible models, since many hyper-parameters should be tuned to particular model
architectures.  

\subsection{Best model approximation}
\begin{figure}[ht]
	\vskip 0.2in
	\begin{center}
		\centerline{\includegraphics[width=\columnwidth]{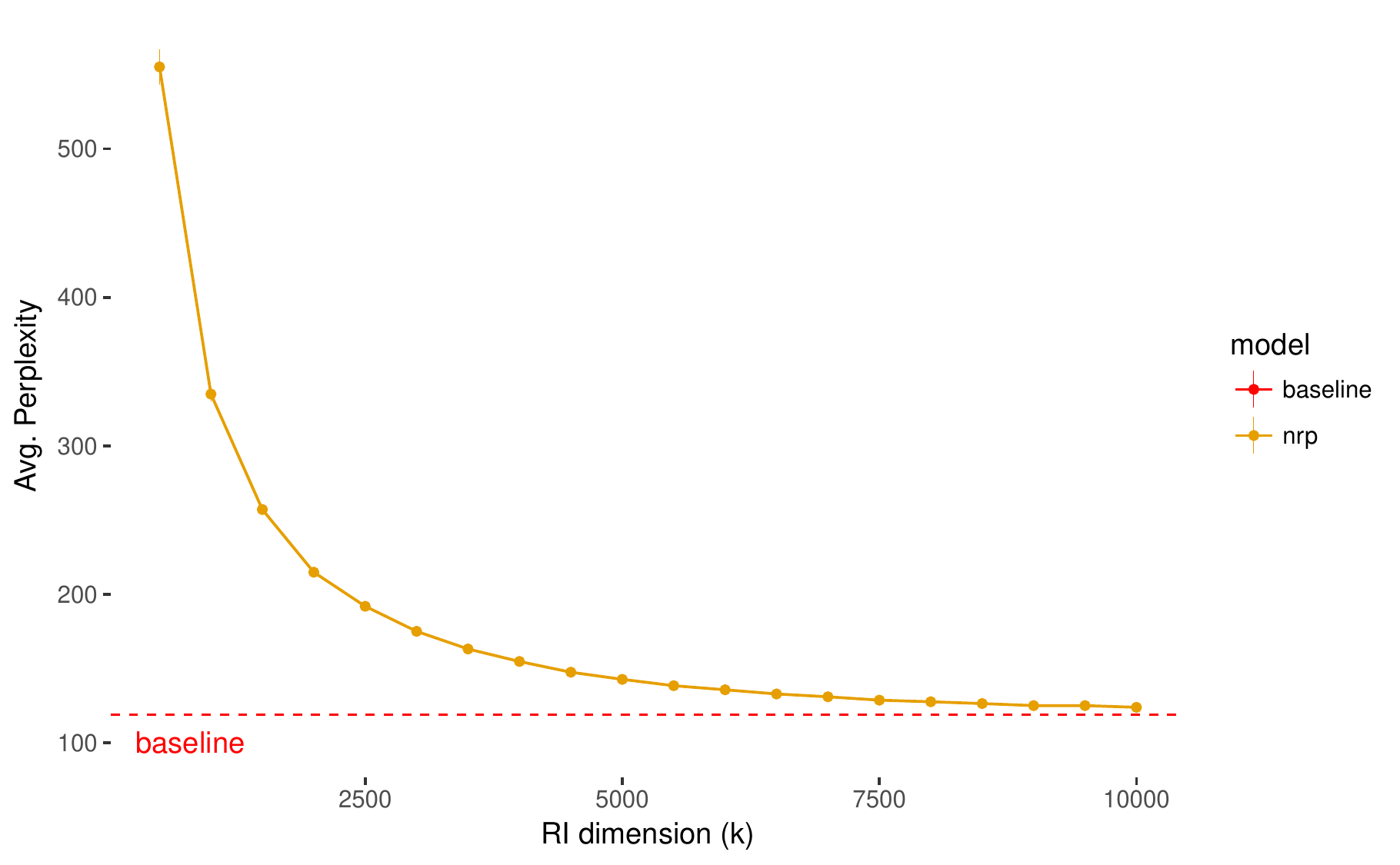}}
		\caption{Average \textit{test} perplexity for the best NRP and baseline configurations 
		found (see tables \ref{tab:best_baseline}, and \ref{tab:best_nrp}) with $m=h=512$ and 
		\textit{dropout} probability $0.6$ for multiple NRP random index vector dimension values $k$.}
		\label{fig:k_curve_nrp_best}
	\end{center}
	\vskip -0.2in
\end{figure}

Finally, we analyse the performance of an NRP model with the best configuration we found in previous experiments, 
and compare it with the best baseline model found. We performed multiple runs with different values of random 
index dimension $k$. The results can be seen in figure \ref{fig:k_curve_nrp_best}.

As with our early experiments, the perplexity decays exponentially with the increase of the random index dimension $k$. 
One key difference is that this decay is much more accentuated and model test perplexity converges to values closer 
to the baseline, with lower-dimensional random indices. In summary, random projections seem to be a viable technique 
to build models with comparable predictive power, with a much lower number of parameters and without the necessity of
\emph{a priori} knowledge of the dimension of the lexicon to be used. Proper regularisation and 
model size tuning was key to achieve good results. NRP models are also more sensitive to hyper-parameter values and 
model size than the baseline model. This was an expected result, since the problem of estimating distributed 
representations using a compressed input space is harder than using dedicated feature vectors for each unique input.

\section{Conclusion and future work}
\label{sec:conclusion}

In this work ,we investigated the potential of using random projections as an encoder for neural network models 
for language modelling. The results extend beyond this sequence prediction task and  can applicable to any 
neural-network-based modelling problem that involves the modelling of discrete input patterns. 

We showed that using random projections allows us to obtain perplexity scores comparable to a model 
that uses an $1\text{-of-}V$ encoding, while reducing the number of necessary trainable parameters. 
On top of this, our approach opens multiple interesting research directions. 

Random projections introduce additional computational complexity in the output layer ,because to obtain 
the output probabilities we need the vector representations for all the existing words --and these 
representations are a combination of multiple features. The problem only arises because we need to 
output a probability distribution over words,. However,our approach is compatible with approximation 
techniques such as \acrfull{nce} \cite{Mnih2012}., with benefits that we will demonstrate in a short while. 
Another future research direction is to answer the question of whether or not more powerful neural network 
architectures (such RNNs or LSTMs) can yield better results using random projections --possibly using 
even smaller random index dimensions and embedding spaces.

We have introduced a new simple baseline on the small, but widely used \textit{Penn Treebank} dataset. 
Our baseline model combined the energy-based principles from the work in \cite{Mnih2007} with the simple 
feedforward seminal neural network architecture proposed in \cite{Bengio2003}, and with a recent 
regularisation technique (dropout \cite{Srivastava2014}). Model performance and training speed 
were also improved by using rectifier linear units (ReLUs) in the hidden layers, similarly to the 
baseline proposed in \cite{Pham2016}. The result was a very simple architecture with a low number of 
parameters, and a performance comparable to similar models in the literature.
We tested our random projection encoder in an energy-based neural network architecture, because this 
would allow us to extend this work beyond the prediction of vocabulary distributions. The fact that we 
can predict distributed representations,means that we can encode --and make predictions about-- 
more complex patterns. In future work, we intend to explore whether or not we can use random 
projections to exploit sequence information, linguistic knowledge, or structured input patterns 
similarly to the approaches in \cite{Plate1995,Cohen2009,Basile2011}.

\bibliographystyle{icml2017}
\bibliography{main}

\end{document}